%% file: main_copy.tex
\documentclass[11pt,a4paper]{article}

\usepackage{float}

\usepackage[utf8]{inputenc}
\usepackage[T1]{fontenc}
\usepackage{microtype}
\usepackage[margin=1in]{geometry}
\usepackage{booktabs}
\usepackage{makecell}
\usepackage{graphicx}
\usepackage{amsmath}
\usepackage{amssymb}
\usepackage{siunitx}
\usepackage{xcolor}
\usepackage{hyperref}
\usepackage[round]{natbib}

\graphicspath{{figures/}}
\hypersetup{
  colorlinks=true,
  linkcolor=blue!50!black,
  citecolor=blue!50!black,
  urlcolor=blue!50!black
}

\newcommand{\probe}{linear-probe accuracy}

\title{\textbf{Predictive Objectives Discard Exogenous\\ Control-Relevant Features: A Controlled Mechanistic Study}}

\author{Ayan Pendharkar}

\date{}

\begin{document}
\maketitle

\begin{abstract}
\noindent
Joint-embedding predictive (JEPA-style) objectives learn representations by predicting future
latents. In doing so, they can discard features that are \emph{exogenous} (uncontrollable by the
agent) yet control-relevant, even when those features are trivially encodable. This occurs because
the objective optimizes temporal predictability rather than control-relevance. We isolate this
failure mode in a controlled $2\times2$ experimental design that varies feature controllability and
relevance independently, using a \emph{predictability knob} that decouples a feature's temporal
predictability from its control-relevance. Comparing six objectives: reconstruction, JEPA,
action-conditioned JEPA, controllability-based JEPA, inverse dynamics under a random policy, and
reward-grounded JEPA,we observe that all evaluated reward-free predictive objectives leave the
\emph{exogenous control-relevant feature} near chance accuracy, while a reward-grounded variant
retains it selectively. The remedy is label-efficient and robust: as little as $2\%$ of
reward-labeled transitions recovers the feature, the effect holds across two environments with
different surface forms, and it persists across latent dimensions from $16$ to $1024$. Comparing the
learned latent geometry against bisimulation theory's prediction, the JEPA latent realizes only a
small fraction of the class separation a supervised reference attains. Our aim is not to
benchmark world models but to isolate an objective-level failure mode under controlled conditions.
\end{abstract}

\section{Introduction}

A representation can fail a downstream task not because it lost something irrelevant, but because
it lost something it could not predict. We study features that are simultaneously reward-relevant,
visually salient (a localized patch in the observation), and temporally unpredictable (their value
next step is close to a coin flip). A self-prediction objective receives no gradient that asks it to
carry such a feature: from the objective's point of view it is noise, even though it determines the
agent's payoff. Predictability and control-relevance, in other words, can be orthogonal, and an
objective that optimizes the former can discard a feature that scores high on the latter.

Joint-embedding predictive architectures predict future latent states rather than reconstructing
pixels~\citep{assran2023ijepa,vjepa2_2025}. Discarding ``unpredictable detail'' is intentional, and
it is what makes these objectives scalable and robust to nuisance texture. The question we
investigate is deliberately narrow: does ``unpredictable'' always coincide with
``irrelevant''? When the two come apart, a representation optimized for predictability throws
away exactly the information a controller needs.

This tension is sharp against bisimulation theory. \citet{zhang2020dbc} argue that an optimal
control representation should keep features that distinguish states by their reward-relevant
futures and collapse everything else; a JEPA objective instead keeps what is predictable. For a
feature that is reward-relevant but temporally unpredictable, the two criteria point in opposite
directions: bisimulation says keep it, self-prediction says drop it. The disagreement is not
hypothetical. It is realized by a single cell of a simple taxonomy.

We organize features along two binary axes. \emph{Controllability} asks whether the agent's action
moves the feature, and \emph{relevance} asks whether the feature sets the reward or the optimal
action. The two axes give four cells. (i) Controllable and relevant: every objective keeps it
(trivial). (ii) Controllable and irrelevant: controllability-based methods waste capacity on it
while bisimulation drops it. (iii) Uncontrollable and irrelevant: the classic exogenous distractor,
studied extensively. (iv) \textbf{Uncontrollable and relevant}: the \emph{exogenous
control-relevant} feature, where the failure mode appears. Prior work has studied cells (i) to (iii)
extensively. Cell~4 has been named as a risk but, to our knowledge, not measured in controlled
isolation.

\paragraph{Contributions.}
\begin{itemize}
  \item \textbf{A controlled isolation of the exogenous control-relevant case.} We provide a
  controlled isolation of cell~4 (exogenous $+$ relevant), to our knowledge not previously measured
  in this isolated form, using a predictability knob that independently varies a feature's temporal
  predictability and its control-relevance. This rules out ``the objective discarded it because it
  was irrelevant'' as an explanation.
  \item \textbf{Empirical measurement of the failure mode and a selective remedy.} The four
  reward-free predictive objectives we evaluate (JEPA and its action-conditioned,
  controllability-based, and inverse-dynamics variants) do not retain cell~4 (\probe{}
  $\approx 0.51$, mutual information $\approx 0$ nats, effective rank $38$ to $42$ confirming the
  representation has \emph{not} collapsed). A reward-grounded variant retains it selectively (probe
  $1.00$, MI $=0.693$ nats, retaining exactly the relevant cells~1 and~4). Reconstruction retains it
  too, but only by indiscriminately keeping every feature, including the irrelevant distractors.
  \item \textbf{Practical scaling findings.} The reward-grounded remedy needs as little as $2\%$ of
  transitions reward-labeled, the effect persists from $16$ to $1024$ latent dimensions, and it
  replicates across two environments with different surface forms.
\end{itemize}

\noindent
Our aim is not to benchmark existing world models, but to isolate an objective-level failure mode
under controlled conditions. Accordingly, all experiments are in small synthetic environments by
design: we seek to demonstrate the \emph{existence} of the failure mode and the mechanism behind it,
not to estimate how often it arises in natural data.

\noindent
Figure~\ref{fig:matrix} (Section~\ref{sec:cell4}) previews the central result: only reward grounding
(and the supervised and reconstruction references) retains the exogenous control-relevant feature,
while all evaluated reward-free predictive objectives leave it near chance.

\section{Background}

\subsection{JEPA and latent self-prediction}
A joint-embedding predictive architecture trains an encoder to predict the latent of a future (or
masked) observation from the latent of the present one, without reconstructing pixels
~\citep{assran2023ijepa}. Because the loss only rewards predicting the part of the future that
\emph{is} predictable, a feature whose future value is unpredictable receives no signal to be
retained. This is a design choice, and it is what lets these models ignore nuisance
detail. It is also central to the program of latent-space world models for autonomous agents
~\citep{lecun2022path}.

\subsection{Bisimulation metrics}
Bisimulation metrics define a distance on states that respects reward and transition structure:
two states with the same reward-relevant future are at distance zero, and states differing in a
reward-relevant feature are at positive distance~\citep{ferns2004metrics}. \citet{zhang2020dbc}
(DBC) showed that representations trained to match a bisimulation metric are substantially more
robust to task-irrelevant distractors than reconstruction- or contrastive-based representations,
because the metric explicitly factors out reward-irrelevant variation. We rely on the converse: a
reward-relevant feature has \emph{positive} bisimulation distance and must be kept.
Our empirical comparison (Section~\ref{sec:obs1}) measures the gap between the analytical
bisimulation distance of the exogenous control-relevant feature and the class separation a trained
JEPA latent actually realizes.

\subsection{A feature taxonomy and the exogenous control-relevant cell}
\label{sec:taxonomy}
We treat the two axes as definitions. A feature is \emph{controllable} if $c_{t+1}$ is a function
of the action $a_t$ (the agent can move it) and \emph{uncontrollable}/\emph{exogenous} if $c_{t+1}$
evolves independently of $a_t$. A feature is \emph{relevant} if it determines the reward, and hence
the optimal action $a^\ast$, and \emph{irrelevant} if it has no effect on reward. The four cells of
the resulting taxonomy are not a new framework but a controlled experimental design that lets us
vary one property at a time. Our focus is cell~4, the uncontrollable-but-relevant case. It is the
hard case for a specific reason: controllability-based methods cannot retain it because the agent
cannot move it, prediction-based methods cannot because the agent cannot predict it, and only
reward-based methods have a signal that distinguishes it from an irrelevant exogenous feature.

\section{Setup}
\label{sec:setup}

\subsection{Environments}
The primary environment, \texttt{QuadrantEnv}, emits $12\times12$ grayscale observations. A single
$3\times3$ corner patch encodes the target feature bit, superimposed on a rich,
deterministically-advancing multi-frequency background (four spatial frequencies with a
deterministic phase progression). The background is therefore \emph{predictable} structure that a
self-prediction encoder can and does model, which is what makes the contrast with the unpredictable
feature patch clean.

We replicate the central result on two surface forms. \texttt{SwitchColorEnv} ($32\times32$
observations) encodes the bit in patch intensity over a sinusoidal grating. The gridworld form uses
a bottom-right patch over a drifting texture with an agent marker. The two forms share the cell-4
logic but differ in resolution, texture statistics, and intrinsic dimensionality, which lets us
separate the objective-level effect from any one rendering.

Feature dynamics follow a single specification. Controllable features toggle on the action taken,
and exogenous features evolve with $P(c_{t+1}=c_t)=p_{\text{repeat}}$, the predictability knob
(default $p_{\text{repeat}}=0.5$, i.e.\ maximally unpredictable). The optimal action is
$a^\ast=c_t$ and the reward is $1$ iff the action matches the relevant feature bit. A helper
\texttt{make\_cell(cell, predictability)} instantiates the single-feature specification for each
quadrant cell. Unless otherwise stated, the main experiments train for $4000$ steps with batch
size $128$, learning rate $10^{-3}$, latent dimension $128$, and three seeds $\{0,1,2\}$
(Appendix~\ref{app:env}).

The comparison across objectives is meaningful only under a set of implementation invariants. Most
importantly, a byte-identical encoder shared by every objective and probes that read the online
encoder on a disjoint evaluation seed, which we enforce and, where possible, assert automatically.
The full list is in Appendix~\ref{app:repro}.

\subsection{The predictability knob}
The knob $p_{\text{repeat}}\in[0.5,1.0]$ controls only predictability. At $p_{\text{repeat}}=0.5$
the feature is an i.i.d.\ Bernoulli draw (maximally unpredictable); at $p_{\text{repeat}}=1.0$ it is
constant (perfectly predictable). Control-relevance is held fixed across the whole
range: the feature always determines the reward. This decoupling is the methodological core of the
study. It lets us ask whether JEPA's information loss is a function of \emph{predictability} rather
than \emph{relevance}, and it rules out ``the objective merely dropped an irrelevant feature'' as an
explanation. All main results are reported at $p_{\text{repeat}}=0.5$. As a sanity check, sweeping
the knob in \texttt{SwitchColorEnv} shows JEPA recovers the feature only as it becomes predictable,
while reconstruction keeps it throughout (Appendix~\ref{app:env}).

\subsection{Objectives}
\label{sec:objectives}
We compare six objectives plus a label-supervised reference (a model trained on ground-truth feature
labels, included as an upper-bound check). The encoder is byte-identical across all of them
(invariant~1); they differ only in which signal the objective consumes (Table~\ref{tab:objectives}).

\begin{table}[H]
\centering
\caption{The six objectives plus a supervised reference. The encoder is byte-identical across all;
each differs only in the signal its objective consumes. \texttt{jepa\_reward} is the proposed
reward-grounded variant; the supervised reference consumes ground-truth feature labels and is an
upper-bound check, not a practical method.}
\label{tab:objectives}
\begin{tabular}{ll>{\raggedright\arraybackslash}p{5.6cm}}
\toprule
\textbf{Objective} & \textbf{Signal consumed} & \textbf{Key property} \\
\midrule
\texttt{recon}        & pixels                                   & pixel-reconstruction reference; keeps everything \\
\texttt{jepa}         & none (latent self-prediction)            & the baseline under study \\
\texttt{jepa\_ac}     & actions taken (predictor on $a_t$)       & action-conditioned; still reward-free \\
\texttt{jepa\_ctrl}   & actions taken (inverse-dynamics head)    & controllability-based; reward-free \\
\texttt{jepa\_invdyn} & actions taken (random policy)            & informative only if the policy correlates with $c$ \\
\texttt{jepa\_reward} & reward labels ($+$ optional bisim term)  & the proposed reward-grounded variant \\
Supervised            & ground-truth feature labels              & label-supervised reference; not a practical method \\
\bottomrule
\end{tabular}
\end{table}

For \texttt{jepa\_invdyn} the policy matters. Under the random-policy stream used in the failure
experiments, the actions carry no information about an exogenous feature, so the inverse-dynamics
head has nothing to anchor cell~4 to and drops it. Under an informative policy whose actions
correlate with the feature (the condition shown in the full matrix, Table~\ref{tab:quadrant_matrix}
in Appendix~\ref{app:full}), the same objective can retain it. But that is ground-truth action
supervision, not unsupervised rescue. We therefore report \texttt{jepa\_invdyn} as a reward-free
objective evaluated under random actions, and flag the informative-action result separately.

\subsection{Metrics}
\label{sec:metrics}
We report three metrics, each chosen to answer a distinct question.
\begin{itemize}
  \item \textbf{Linear probe accuracy} (primary): can the feature $c$ be decoded by a linear
  classifier on the frozen latent $z$? Chance is $0.50$ and we use a retain threshold of $0.75$.
  \item \textbf{InfoNCE mutual information}: a lower-bound estimate of $I(Z;c)$ in nats trained on a
  held-out split, with theoretical maximum $\log 2\approx0.693$ for a one-bit feature
  ~\citep{oord2018cpc}. It corroborates the probe without sharing its decision boundary.
  \item \textbf{Effective rank}: $\exp$ of the entropy of the normalized singular values of the
  latent matrix, reported alongside every probe so that we can separate \emph{selective
  dropping} (high rank, the feature is gone but the representation is rich) from
  \emph{representational collapse} (low rank, everything is gone). We describe a result as
  selective dropping only when effective rank exceeds $5$.
\end{itemize}

\noindent
Because the environments are deterministic apart from seed initialization, and because the observed
effect sizes are large (retention is close to either chance or ceiling rather than intermediate), we
report means and standard deviations over three seeds rather than running a larger seed sweep.

\section{Results}

\subsection{Reward-free predictive objectives do not retain the exogenous control-relevant feature}
\label{sec:cell4}

\begin{figure}[H]
\centering
\includegraphics[width=0.64\linewidth]{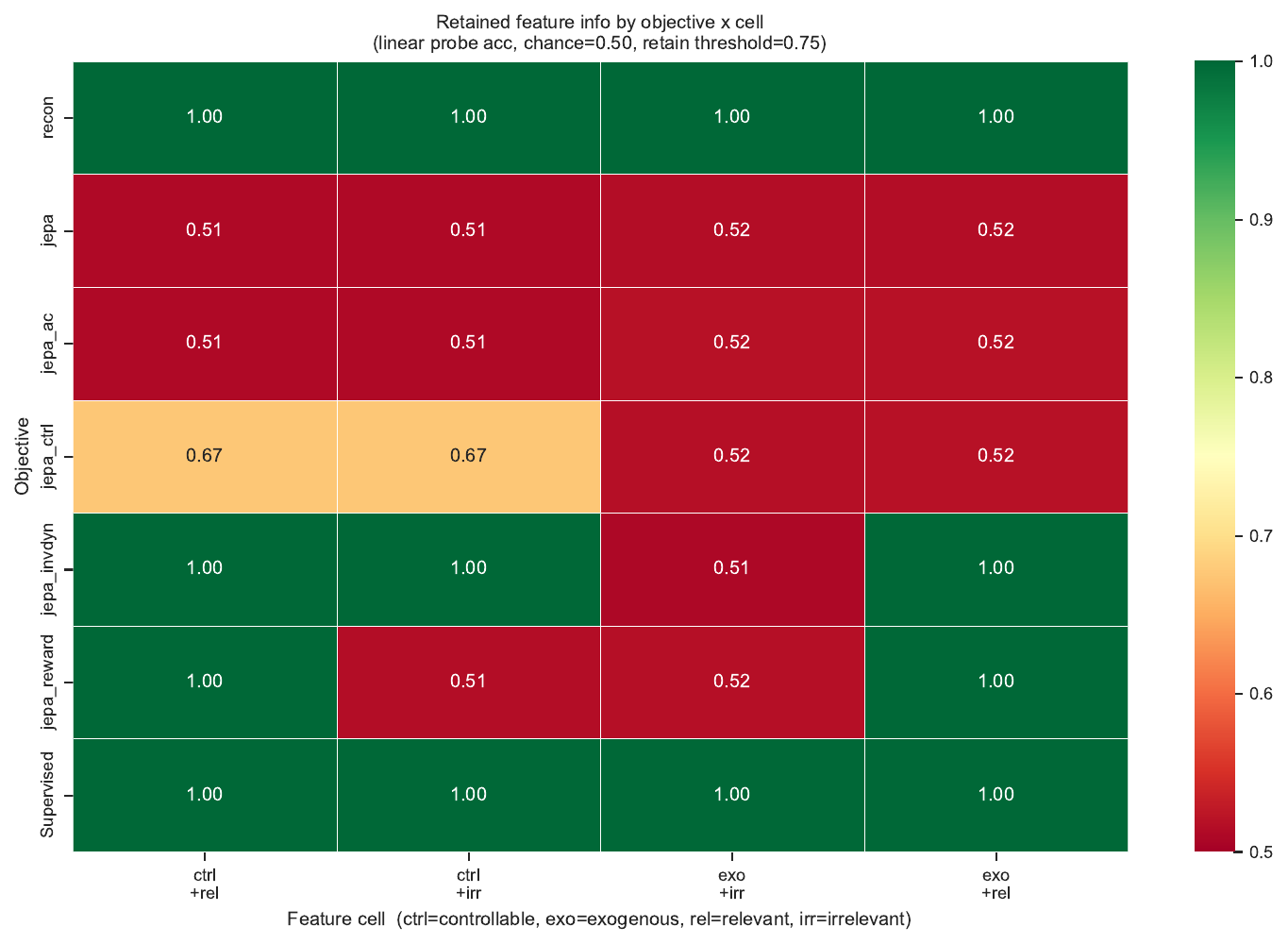}
\caption{\textbf{Objective $\times$ cell retention matrix} (\probe{} on \texttt{QuadrantEnv}; green
$\geq0.75$ retained, red $<0.75$ not retained). The exogenous control-relevant column (cell~4) is
the paper's central result: only the reward-grounded variant and the references retain the feature.
\texttt{jepa\_reward} retains exactly the relevant cells (1 and 4) and not the irrelevant ones (2
and 3), whereas \texttt{jepa\_ctrl} partially recovers the controllable cells (1--2, $\approx0.67$)
and not the exogenous ones; all purely self-predictive objectives retain nothing under
random-policy data.}
\label{fig:matrix}
\end{figure}

The objective $\times$ cell retention matrix (Figure~\ref{fig:matrix}, numerical values in
Table~\ref{tab:quadrant_matrix}, Appendix~\ref{app:full}) is the central summary of this study. The
exogenous control-relevant column (cell~4) constitutes the central empirical result of this paper:
across the seven rows, the only objectives that retain it are \texttt{jepa\_reward} and the two
references (\texttt{recon} and the supervised reference). The rest of the matrix shows \emph{why}
this matters: reward grounding's selectivity follows \emph{relevance} (it retains cells~1 and~4, not
2 or 3), whereas the controllability-based objective's follows \emph{controllability} (cells~1
and~2). We now examine cell~4 in detail across both environments.

\begin{figure}[H]
\centering
\includegraphics[width=0.95\linewidth]{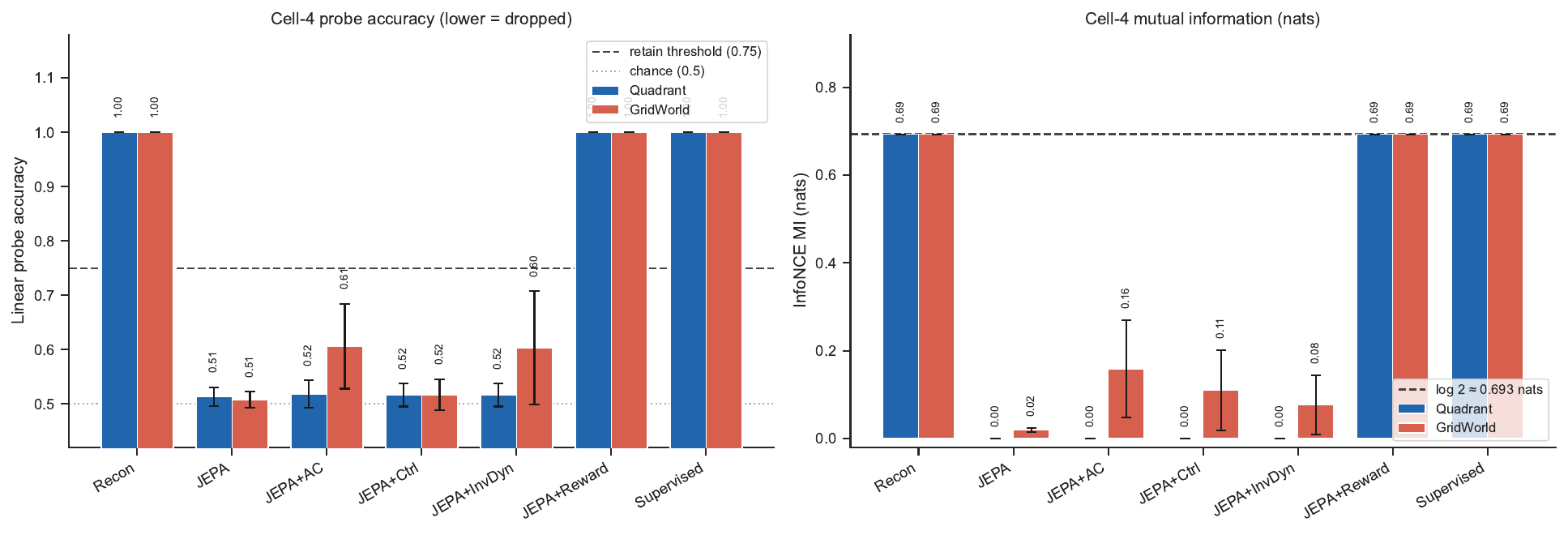}
\caption{\textbf{Cell-4 results across both environments.} Left: \probe{} (chance $=0.50$, retain
threshold $=0.75$) for each objective on the exogenous control-relevant feature. Right: InfoNCE
mutual information (nats; $\log 2\approx0.693$ is the maximum for a one-bit feature). All
evaluated reward-free predictive objectives, including action-conditioned JEPA, controllability-based
JEPA, and inverse dynamics with random actions, leave the feature near chance (probe $\approx0.5$,
MI $\approx 0$). The reward-grounded variant recovers it at the theoretical ceiling, matching the
\texttt{recon} and supervised references. Bars are mean $\pm$ std over three seeds.}
\label{fig:cell4}
\end{figure}

All evaluated reward-free predictive objectives leave the exogenous control-relevant feature
near chance, in both environments and across three seeds (Figure~\ref{fig:cell4},
Table~\ref{tab:cell4_failure}). On \texttt{QuadrantEnv}, \texttt{jepa}
reaches \probe{} $0.51\pm0.02$ with mutual information $0.00\pm0.00$ nats; \texttt{jepa\_ac}
$0.52\pm0.03$; \texttt{jepa\_ctrl} $0.52\pm0.02$; and \texttt{jepa\_invdyn} (random policy)
$0.52\pm0.02$, all at chance with essentially zero retained information. In contrast,
\texttt{jepa\_reward} reaches $1.00\pm0.00$ with $0.69\pm0.00$ nats, the one-bit ceiling, matching
the \texttt{recon} and supervised references.

\input{table2_cell4}

The picture is the same on \texttt{GridWorldHiddenRuleEnv}, with two informative wrinkles. First,
the order is preserved: \texttt{jepa} sits at $0.51\pm0.02$ and \texttt{jepa\_reward} at
$1.00\pm0.00$. Second, the action-using objectives are noticeably noisier on this richer surface
form: \texttt{jepa\_ac} reaches $0.61\pm0.08$ and \texttt{jepa\_invdyn} $0.60\pm0.10$
(Table~\ref{tab:cell4_failure}). The large standard deviations are themselves the point. On some
seeds the random actions happened to carry marginal information about the rule bit, which is
exactly what the \texttt{jepa\_invdyn} caveat predicts: its retention tracks action
informativeness rather than the feature itself. We therefore report these as mean $\pm$ std and
never as a mean alone.

\paragraph{The feature is selectively omitted rather than lost through representational collapse.}
On \texttt{QuadrantEnv} every objective that
discards it retains a high effective rank, between $38$ and $42$; the representation is rich, it has
simply not encoded this one feature. On \texttt{GridWorldHiddenRuleEnv} the discarding objectives
have effective rank near $3$ (range $2.75$ to $4.61$ across all cell-4 runs), but this reflects the
intrinsic dimensionality of the simpler images rather than collapse: \texttt{recon} retains the
feature perfectly at effective rank $3.30$. ``Low-rank environment'' and ``collapsed
representation'' are distinct, and only the latter would undercut the claim. Here the environment is
simply low-dimensional while the feature is selectively omitted.

\subsection{Reward grounding is selectively relevant, not selectively controllable}
\label{sec:selective}

The reward-grounded variant does not merely retain cell~4; it retains the right \emph{axis}.
\texttt{jepa\_reward} keeps both relevant cells, cell~1 (controllable) and cell~4 (exogenous), and
does not retain either irrelevant cell, cell~2 or cell~3 (Figure~\ref{fig:matrix},
Table~\ref{tab:quadrant_matrix}).
Its mutual information is $0.693$ nats exactly where the probe is $1.00$ and $0$ nats where
the probe is $\approx0.51$, with zero variance across seeds. Its selectivity axis is relevance.

Contrast this with the controllability-based objective. \texttt{jepa\_ctrl} keeps cells~1 and~2
(both \emph{controllable}, relevant or not, at $\approx0.67$) and drops cells~3 and~4 (both
exogenous). Its selectivity axis is controllability, which is the wrong axis for control
performance: it spends capacity on a controllable-irrelevant feature while discarding the
exogenous-relevant one. The distinction between controllability-based and reward-based selectivity
is not cosmetic. \texttt{jepa\_ctrl} does not retain the exogenous control-relevant feature while
spending capacity on the controllable-irrelevant one. (We note that the \texttt{jepa\_ctrl} cell-1/2
value of $0.67\pm0.23$ is driven by a single seed reaching $1.00$ while the others stay near chance,
so it is a partial and unstable recovery, not a clean retain; Table~\ref{tab:quadrant_matrix}.)

\subsection{The failure is not a capacity problem}
\label{sec:capacity}

\begin{figure}[H]
\centering
\includegraphics[width=0.7\linewidth]{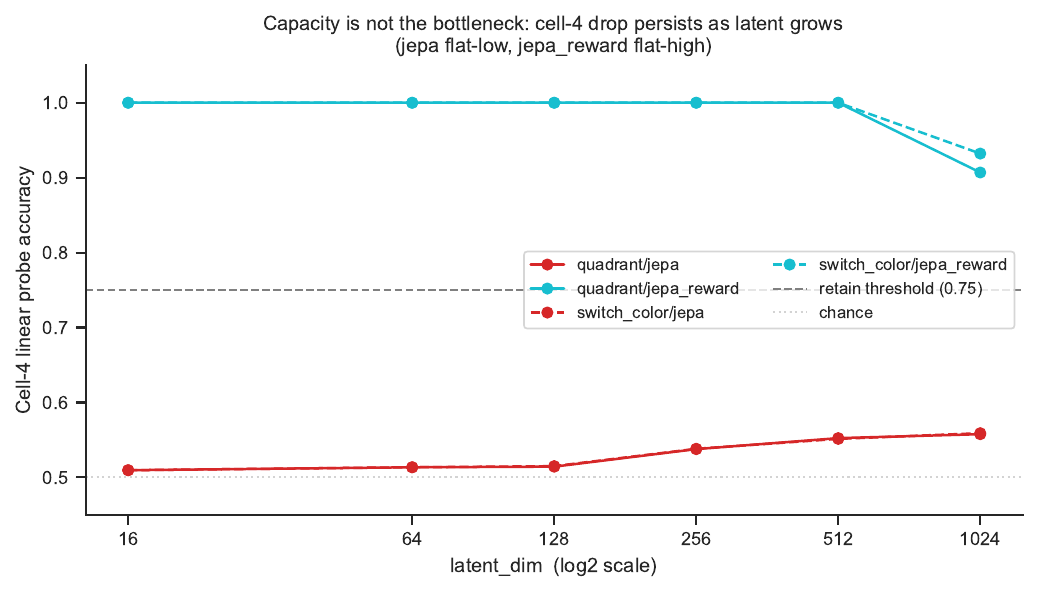}
\caption{\textbf{Cell-4 retention versus latent dimension} for \texttt{jepa} and
\texttt{jepa\_reward} on \texttt{QuadrantEnv} and \texttt{SwitchColor}. JEPA stays near chance
across every latent dimension from $16$ to $1024$, never approaching the retain threshold.
\texttt{jepa\_reward} stays near perfect from $16$ to $512$, with a slight dip at $1024$
attributable to training budget at that scale. The failure is objective-structural, not
architectural.}
\label{fig:capacity}
\end{figure}

Increasing latent capacity does not rescue the feature for JEPA. As the latent
dimension grows from $16$ to $1024$, JEPA's cell-4 \probe{} moves only from $0.510$ to $0.558$ on
\texttt{QuadrantEnv} and never crosses the $0.75$ retain threshold (Figure~\ref{fig:capacity},
Table~\ref{tab:capacity_minreward}). The slight upward drift at very large dimensions is small and
plausibly reflects marginal saturation of the covariance regularizer rather than genuine retention.
\texttt{jepa\_reward} holds at $1.00$ from dimension $16$ through $512$ and dips to $0.907$
(\texttt{Quadrant}) or $0.932$ (\texttt{SwitchColor}) only at $1024$, a training-budget artifact at
the largest scale rather than a capacity floor. The cell-4 failure to retain is a property of the
objective, not of the encoder's size.

\subsection{Reward grounding is label-efficient}
\label{sec:labeleff}

\begin{figure}[H]
\centering
\includegraphics[width=0.7\linewidth]{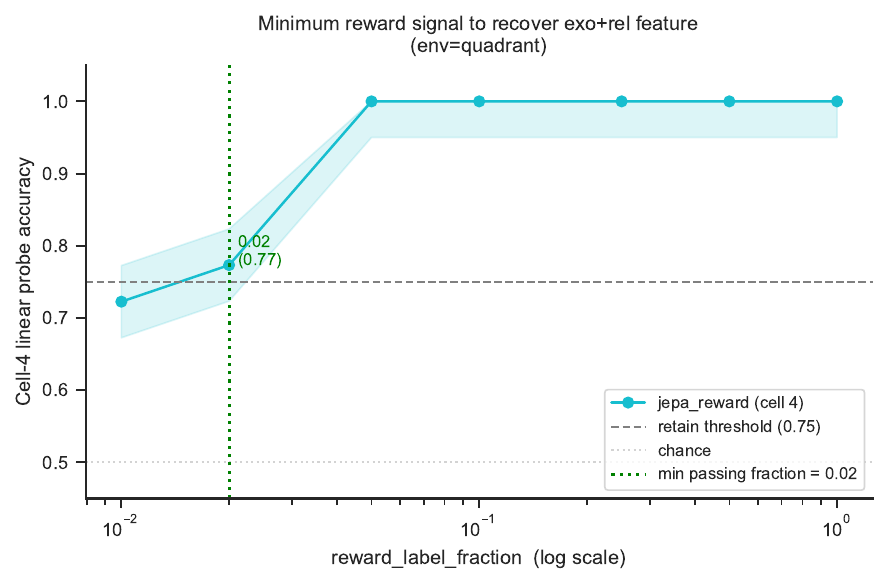}
\caption{\textbf{Cell-4 retention versus the fraction of reward-labeled transitions} (log scale,
\texttt{QuadrantEnv}). The retain threshold ($0.75$) is first crossed at a reward-label fraction of
$0.02$, two percent of transitions (mean accuracy $0.77$). Even $1\%$ of labeled transitions
yields \probe{} $0.72$, already well above the chance level of $0.50$. The reward signal is highly
efficient for recovering the exogenous control-relevant feature.}
\label{fig:minreward}
\end{figure}

The reward signal is also cheap. The smallest reward-label fraction for which \texttt{jepa\_reward}
crosses the retain threshold is $0.02$, one labeled transition in fifty, at mean accuracy $0.77$
(Figure~\ref{fig:minreward}, Table~\ref{tab:capacity_minreward}). At a fraction of $0.01$ (one in a
hundred) the probe is already $0.72$, substantially above the chance level of $0.50$. This
efficiency suggests that relatively sparse reward signals can suffice in our controlled setting.

\section{Empirical Comparison with Bisimulation Predictions}
\label{sec:obs1}

We compare the latent geometry induced by JEPA against the class separation that bisimulation theory
predicts for the exogenous control-relevant cell. We quantify class separation as the whitened
(Mahalanobis) distance between the $c=0$ and $c=1$ latent centroids on a held-out evaluation stream;
a value near zero means the feature is not linearly recoverable from the latent. At convergence on
cell~4 ($p_{\text{repeat}}=0.5$, full training), the analytical on-policy bisimulation distance
between the two classes is $1.0$, and both the supervised reference and the
reconstruction baseline realize a latent class separation of $\approx 1.998$. The JEPA latent
realizes substantially less class separation than both references ($0.105$ versus $\approx 1.998$),
and its \probe{} sits at chance ($0.49$). The gap between the reference separation and the JEPA
separation is $1.893$, which we call the observed separation gap for this comparison.

\begin{figure}[H]
\centering
\includegraphics[width=0.55\linewidth]{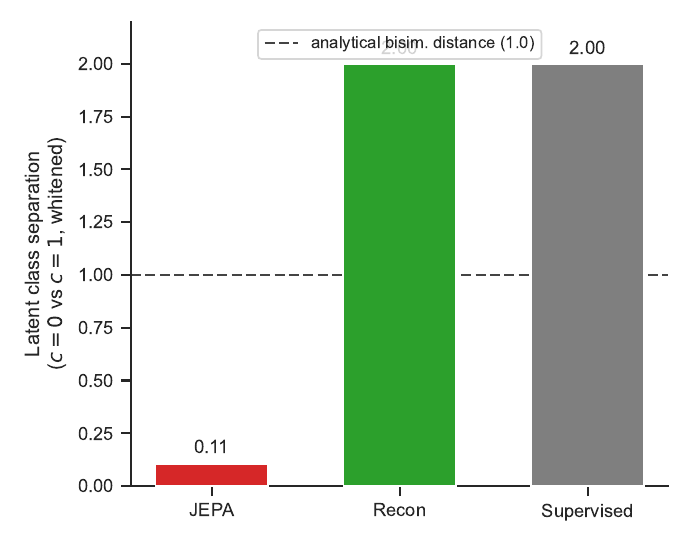}
\caption{\textbf{Class separation} (whitened centroid distance between $c=0$ and $c=1$ latent
representations) for \texttt{jepa}, \texttt{recon}, and the supervised reference at convergence (cell~4,
$p_{\text{repeat}}=0.5$, full training). The horizontal reference marks the analytical bisimulation
distance ($1.0$). JEPA realizes substantially less class separation than the references ($0.105$
versus $\approx1.998$); the separation gap is $1.893$.}
\label{fig:bisim}
\end{figure}

The cell-4 feature has analytical on-policy bisimulation distance $1.0>0$: the immediate-reward gap
is $1.0$ and the discounted exogenous-transition term vanishes at $p_{\text{repeat}}=0.5$, so a
bisimulation-aligned representation would preserve this distinction. Only the supervised reference and the
pixel-grounded reconstruction baseline realize that separation in latent space (class distance
$\approx1.998$); the JEPA latent realizes only $0.105$ (Figure~\ref{fig:bisim}). In our setting,
latent self-prediction provides no effective optimization signal toward the
unpredictable-but-relevant feature and does not encode it, even though it is trivially encodable. These numbers are a single-seed point estimate under full
training and are reported as such.

\citet{zhang2020dbc} relate bisimulation-aligned representations to lower control regret: a
representation that fails to separate states differing in a reward-relevant feature cannot support
a controller that acts on it. This comparison locates exactly such a failure: the near-zero class
separation is measured directly from the latent geometry, so the loss of control-relevant structure
is intrinsic to the representation rather than an artifact of any particular probe or downstream
policy.

We emphasize that this comparison is an empirical measurement in a controlled synthetic environment,
not a formal proof; the analytical bisimulation distance is a closed-form derivation, but the claim
that JEPA does not encode the feature is measured from trained models on a finite budget, and its
generality to large pretrained JEPA models is an open question.

\section{Related Work and Discussion}

\subsection{JEPA and world models}
I-JEPA~\citep{assran2023ijepa}, V-JEPA 2 and V-JEPA 2-AC~\citep{vjepa2_2025}, and
DINO-WM~\citep{zhou2024dinowm} establish the latent-prediction objective and demonstrate its
empirical benefits for representation learning and planning on pre-trained visual features. These
works motivate the objective; our work is complementary, asking what information is structurally at
risk under it.

\subsection{Bisimulation-based representation learning}
DBC~\citep{zhang2020dbc} established that bisimulation-aligned representations are far more robust to
task-irrelevant distractors than reconstruction, by explicitly factoring out reward-irrelevant
variation. DeepMDP~\citep{gelada2019deepmdp} learns latent models with reward and transition losses
that provably bound value error, and action-bisimulation~\citep{actionbisim2024} extends
bisimulation to action-conditioned settings. These lines established that the reward signal shapes
which structure a representation keeps. Our contribution is the controlled isolation of the specific
cell where self-prediction and bisimulation conflict, exogenous and relevant, and a quantitative
measurement of the information gap.

\subsection{Closest prior work}
\paragraph{Bisimulation on JEPA world models (arXiv:2602.18639).}
\citet{bisimjepa2026} apply a bisimulation objective to JEPA world models to handle distractors. We
differ in the cell we target: their distractors are exogenous and \emph{irrelevant} (cell~3), the
classic distractor the agent neither controls nor needs, whereas we target the exogenous and
\emph{relevant} cell~4. Their bisimulation grounding and our reward grounding are related fixes, but
they are motivated by different failure modes.

\paragraph{Sensorimotor world models (arXiv:2606.20104).}
\citet{sensorimotor2026} add an inverse-dynamics regularizer to retain action-relevant features.
Our \texttt{jepa\_invdyn} (random-policy) experiment shows directly that when actions do not encode
the exogenous-relevant feature, inverse dynamics fails: cell-4 retention is only $0.52\pm0.02$ on
\texttt{QuadrantEnv} (Table~\ref{tab:cell4_failure}). That fix works for controllable features but
does not generalize to exogenous-relevant ones without an informative policy.

\paragraph{A critique of world models (arXiv:2507.05169).}
\citet{xing2025critique} name task-relevant information loss as a concern for JEPA-style models. We
make that concern concrete by measuring it, through mutual information, latent class separation, and
probe accuracy, in controlled isolation, and by exhibiting a selective remedy.

\subsection{Why this is not a distractor problem}
A natural objection is that DBC and its successors already showed reward-free objectives mishandle
exogenous features. The distinction is the relevance axis. The classic distractor is exogenous
\emph{and irrelevant} (cell~3): the agent neither controls it nor needs it, and the desired behavior
is to discard it---which reconstruction notably fails to do and which bisimulation handles well. Our
case is exogenous \emph{and relevant} (cell~4): the agent cannot control or predict the feature, yet
the optimal action depends on it, so the desired behavior is to \emph{keep} it. Prior distractor
results therefore do not cover cell~4. In fact the controllability- and prediction-based fixes that
suppress cell-3 distractors are precisely the ones that also suppress the cell-4 feature
(Figure~\ref{fig:matrix}). The relevance axis separates the two cases, and it is the core of our
contribution.

\subsection{Why the result is not trivial}
One might object that it is obvious a predictive objective ignores unpredictable things. The
non-trivial observation is that unpredictability and control-relevance can be \emph{orthogonal}: a
feature can be decisive for the optimal action while being statistically independent of its own
past. When that happens, ``ignore the unpredictable'' and ``keep the control-relevant'' give
opposite instructions, and a purely predictive objective follows the former at the cost of the
latter. Our predictability knob makes this orthogonality explicit and measurable, and the quadrant
matrix shows the resulting loss is selective (the feature is gone while effective rank stays high),
not a generic degradation.

\subsection{What this result does not imply}
To be explicit about scope, our findings do \emph{not} show that V-JEPA or any specific pretrained
model fails on a real task; they do \emph{not} show that all world models or all predictive
objectives are flawed; and they do \emph{not} show that predictive objectives are not useful (they
are demonstrably effective at the representation-learning goals they were designed for). We show only
that, under controlled conditions, an identifiable class of features (exogenous, control-relevant,
and unpredictable) is at structural risk under the reward-free predictive objectives we evaluate,
and that a small amount of reward signal removes that risk in our setting.

\subsection{Limitations}
\label{sec:limits}
We are deliberately explicit about scope: this is a controlled mechanistic study, not a benchmark
and not a theorem.
\begin{itemize}
  \item All results are in small synthetic environments ($12\times12$ and $32\times32$
  observations). We run \emph{no} large-scale pretrained-model experiments (e.g.\ V-JEPA 2-AC),
  \emph{no} robotics experiments, \emph{no} Atari or other standard RL-benchmark experiments, and
  \emph{no} real-world demonstrations; transfer to any of these is untested and is future work.
  \item This is not a benchmark study. We do not rank methods on a task suite and make no claim
  about how frequently the failure mode arises in natural data, only that it exists and has an
  identifiable cause.
  \item The bisimulation comparison (Section~\ref{sec:obs1}) is an empirical measurement, not a
  proved theorem, and is reported from a single seed at full training; a theorem would require
  explicit assumptions about the learning dynamics and the environment class.
  \item The gridworld environment has intrinsically low-rank observations (effective rank
  $\approx3$); the effective-rank anti-collapse check is calibrated for the quadrant environment and
  should be interpreted accordingly there.
  \item The \texttt{jepa\_invdyn} result depends on the action policy: the cell-4 failure holds
  under random actions but need not hold under every informative policy.
  \item Our objective set spans reconstruction, prediction, action-conditioning, controllability,
  inverse dynamics, and reward grounding, but is not exhaustive; we claim the failure for the
  reward-free predictive objectives we evaluate, not for every conceivable objective.
\end{itemize}

\subsection{Future work}
The natural next step is to demonstrate the failure mode on a pretrained V-JEPA 2-AC or DINO-WM
encoder in a goal-not-given task. The minimal-reward-signal result ($2\%$ of transitions) suggests
that even a weakly reward-conditioned adapter on a frozen JEPA encoder may be enough to prevent it.

\subsection{Summary}
Our primary contribution is the controlled isolation and measurement of the exogenous
control-relevant failure mode: a feature that is reward-relevant but temporally unpredictable is
not retained by the reward-free predictive objectives we evaluate, is retained selectively by a
reward-grounded variant, and the effect is selective (high effective rank), label-efficient, and
stable across two environments and latent dimensions from $16$ to $1024$. We make no claim beyond
this controlled setting; whether the same failure mode appears in large pretrained models on natural
tasks is the open question we leave to future work.

\appendix
\section{Environment implementation details}
\label{app:env}
\texttt{QuadrantEnv} renders $12\times12$ grayscale frames: a $3\times3$ corner patch carries the
feature bit, and the background is a sum of four spatial-frequency gratings whose phase advances
deterministically each step, giving the encoder predictable structure to model. \texttt{FeatureSpec}
records, per feature, whether it is controllable (toggled by the action) and whether it is relevant
(sets the reward), and \texttt{make\_cell(cell, predictability)} assembles the single-feature spec
for each quadrant cell. Exogenous features evolve with $P(c_{t+1}=c_t)=p_{\text{repeat}}$; the
optimal action is $a^\ast=c_t$ and the reward is $1$ iff the action matches the relevant bit. The
main experiments use $20000$ training transitions, $4000$ evaluation transitions, batch size $128$,
learning rate $10^{-3}$, latent dimension $128$, $4000$ training steps, and seeds $\{0,1,2\}$. The
two replication surface forms are \texttt{SwitchColorEnv} ($32\times32$, intensity patch over a
sinusoidal grating) and \texttt{GridWorldHiddenRuleEnv} (bottom-right patch over a drifting texture
with an agent marker). As a check on the predictability knob in \texttt{SwitchColorEnv}, JEPA
recovers the feature at $1.00\pm0.00$ when $p_{\text{repeat}}=1.0$ and only $0.61\pm0.13$ when
$p_{\text{repeat}}=0.5$, while reconstruction holds at $1.00\pm0.00$ throughout.

\section{InfoNCE estimator details}
\label{app:infonce}
The mutual-information estimate is a separable InfoNCE critic (a small MLP) trained on a held-out
split so the bound stays honest~\citep{oord2018cpc}, run for $300$ epochs. As a sanity check, the
estimator returns approximately $\log 2\approx0.69$ nats when the feature is perfectly recoverable
and approximately $0$ nats when the feature is independent of the latent, matching the values
observed for \texttt{recon} and the supervised reference and for collapsed JEPA latents respectively
(Table~\ref{tab:cell4_failure}).

\section{Full results}
\label{app:full}
Table~\ref{tab:quadrant_matrix} gives the full objective $\times$ cell matrix on
\texttt{QuadrantEnv} (mean $\pm$ std over three seeds). In this matrix \texttt{jepa\_invdyn} is
trained on informative-action data ($a^\ast$ with $\varepsilon=0.2$ exploration); its cell-4
retention of $1.00$ there tracks action informativeness, not reward grounding, and under random
actions it drops to chance (Table~\ref{tab:cell4_failure}). Table~\ref{tab:capacity_minreward} gives
the capacity sweep across both environments and the minimum reward-label fraction.

\input{table1_quadrant}

\input{table3_capacity_minreward}

\section{Implementation invariants and reproducibility}
\label{app:repro}
The comparison across objectives is meaningful only under a set of implementation invariants, which
we enforce throughout and, where possible, assert automatically.
\begin{enumerate}
  \item \textbf{Byte-identical encoder.} The encoder architecture is identical (matching
  \texttt{state\_dict} keys and shapes) across every objective, so any difference in retention is
  attributable to the objective, not the encoder. This is asserted automatically.
  \item \textbf{Anti-collapse regularization.} JEPA objectives use VICReg-style variance (and
  covariance) regularization~\citep{bardes2022vicreg}; we confirm the latent has not collapsed by
  reporting effective rank alongside every probe (Section~\ref{sec:metrics}).
  \item \textbf{BYOL-style target.} The prediction target is produced by an exponential
  moving-average (EMA) copy of the online encoder~\citep{grill2020byol} (decay $0.996$) with a
  normalized prediction loss.
  \item \textbf{Honest probing.} All probes read the \emph{online} encoder (never the EMA target) on
  a disjoint evaluation seed; this invariant is asserted automatically.
  \item \textbf{Configurable capacity and idempotent caching.} The latent dimension is configurable
  (here $128$, swept $16$--$1024$ in the capacity study) and ``skip-if-done'' is keyed on result
  JSONs rather than checkpoints.
\end{enumerate}
Results are deterministic given fixed seeds $\{0,1,2\}$ and were produced on a GPU (CUDA, PyTorch).
An automated check (\texttt{verify.py}) asserts invariants~1 and~4 and exercises the InfoNCE sanity
check above. Method names in the tables and figures match the released code, except that the
label-supervised reference appears there under the identifier \texttt{oracle}. Full reproduction
instructions and the per-run result JSONs that back every number in this paper are described in the
accompanying \texttt{REPRODUCE.md}.

\bibliographystyle{plainnat}
\bibliography{references}

\end{document}

%% file: table2_cell4.tex
\begin{table}[H]
\centering
\caption{Cell-4 failure across both environments (mean\,$\pm$\,std, 3 seeds, random-action policy). Probe acc\,$<$\,0.75 = feature dropped. MI in nats; log\,2\,$\approx$\,0.693 = fully retained. Note large std for \texttt{jepa\_invdyn}: its rescue tracks action informativeness rather than the feature itself.}
\label{tab:cell4_failure}
\begin{tabular}{lcccc}
\toprule
 & \multicolumn{2}{c}{\textbf{Quadrant}} & \multicolumn{2}{c}{\textbf{GridWorld}} \\
\cmidrule(lr){2-3}\cmidrule(lr){4-5}
\textbf{Objective} & Probe acc & MI (nats) & Probe acc & MI (nats) \\
\midrule
  \texttt{recon} & $1.00 \pm 0.00$ & $0.69 \pm 0.00$ & $1.00 \pm 0.00$ & $0.69 \pm 0.00$ \\
  \texttt{jepa} & $0.51 \pm 0.02$ & $0.00 \pm 0.00$ & $0.51 \pm 0.02$ & $0.02 \pm 0.00$ \\
  \texttt{jepa\_ac} & $0.52 \pm 0.03$ & $0.00 \pm 0.00$ & $0.61 \pm 0.08$ & $0.16 \pm 0.11$ \\
  \texttt{jepa\_ctrl} & $0.52 \pm 0.02$ & $0.00 \pm 0.00$ & $0.52 \pm 0.03$ & $0.11 \pm 0.09$ \\
  \texttt{jepa\_invdyn} & $0.52 \pm 0.02$ & $0.00 \pm 0.00$ & $0.60 \pm 0.10$ & $0.08 \pm 0.07$ \\
  \texttt{jepa\_reward} & $1.00 \pm 0.00$ & $0.69 \pm 0.00$ & $1.00 \pm 0.00$ & $0.69 \pm 0.00$ \\
  Supervised & $1.00 \pm 0.00$ & $0.69 \pm 0.00$ & $1.00 \pm 0.00$ & $0.69 \pm 0.00$ \\
\bottomrule
\end{tabular}
\end{table}

%% file: table1_quadrant.tex
\begin{table}[H]
\centering
\caption{Full objective $\times$ cell retention matrix. Values are linear probe accuracy (mean\,$\pm$\,std over 3 seeds, chance\,=\,0.50). Highlighted: cell 4 (exo\,+\,rel), the hard case where only reward-grounded signals retain the feature.}
\label{tab:quadrant_matrix}
\begin{tabular}{lcccc}
\toprule
\textbf{Objective} & \makecell{\textbf{Cell 1}\\\\\textbf{ctrl+rel}} & \makecell{\textbf{Cell 2}\\\\\textbf{ctrl+irr}} & \makecell{\textbf{Cell 3}\\\\\textbf{exo+irr}} & \makecell{\textbf{Cell 4}\\\\\textbf{exo+rel}} \\
\midrule
  \texttt{recon} & $1.00 \pm 0.00$ & $1.00 \pm 0.00$ & $1.00 \pm 0.00$ & $1.00 \pm 0.00$ \\
  \texttt{jepa} & $0.51 \pm 0.01$ & $0.51 \pm 0.01$ & $0.52 \pm 0.02$ & $0.52 \pm 0.02$ \\
  \texttt{jepa\_ac} & $0.51 \pm 0.01$ & $0.51 \pm 0.01$ & $0.52 \pm 0.03$ & $0.52 \pm 0.03$ \\
  \texttt{jepa\_ctrl} & $0.67 \pm 0.23$ & $0.67 \pm 0.23$ & $0.52 \pm 0.02$ & $0.52 \pm 0.02$ \\
  \texttt{jepa\_invdyn} & $1.00 \pm 0.00$ & $1.00 \pm 0.00$ & $0.51 \pm 0.03$ & $1.00 \pm 0.00$ \\
  \texttt{jepa\_reward} & $1.00 \pm 0.00$ & $0.51 \pm 0.01$ & $0.52 \pm 0.02$ & $1.00 \pm 0.00$ \\
  Supervised & $1.00 \pm 0.00$ & $1.00 \pm 0.00$ & $1.00 \pm 0.00$ & $1.00 \pm 0.00$ \\
\bottomrule
\end{tabular}
\end{table}

%% file: table3_capacity_minreward.tex
\begin{table}[H]
\centering
\caption{(Top) Capacity sweep: cell-4 linear probe accuracy vs latent dimension for \texttt{jepa} and \texttt{jepa\_reward} on both environments (3 seeds; values are means where per-seed data is available, else mean only). \texttt{jepa} stays near chance and \texttt{jepa\_reward} stays near perfect across all sizes, ruling out capacity as the bottleneck. (Bottom) Minimum reward-signal fraction: the smallest \texttt{reward\_label\_fraction} for which \texttt{jepa\_reward} retains cell 4 (probe acc\,$\geq$\,0.75) on Quadrant (3 seeds, mean shown).}
\label{tab:capacity_minreward}
\begin{tabular}{lcccc}
\toprule
\textbf{latent\_dim} & \multicolumn{2}{c}{\textbf{Quadrant}} & \multicolumn{2}{c}{\textbf{SwitchColor}} \\
\cmidrule(lr){2-3}\cmidrule(lr){4-5}
 & \texttt{jepa} & \texttt{jepa\_reward} & \texttt{jepa} & \texttt{jepa\_reward} \\
\midrule
  16 & $0.51$ & $1.00$ & $0.51$ & $1.00$ \\
  64 & $0.51$ & $1.00$ & $0.51$ & $1.00$ \\
  128 & $0.51$ & $1.00$ & $0.52$ & $1.00$ \\
  256 & $0.54$ & $1.00$ & $0.54$ & $1.00$ \\
  512 & $0.55$ & $1.00$ & $0.55$ & $1.00$ \\
  1024 & $0.56$ & $0.91$ & $0.56$ & $0.93$ \\
\midrule
\multicolumn{5}{l}{\textit{Minimum reward-label fraction (Quadrant, cell 4)}} \\
\midrule
\textbf{reward\_frac} & \multicolumn{4}{c}{\textbf{Cell-4 probe acc (mean)}} \\
\midrule
  0.01 & $0.72$ &  \\
  0.02 & $0.77$ &  $\leftarrow$ min passing \\
  0.05 & $1.00$ &  \\
  0.1 & $1.00$ &  \\
  0.25 & $1.00$ &  \\
  0.5 & $1.00$ &  \\
  1 & $1.00$ &  \\
\bottomrule
\end{tabular}
\end{table}